\pdfoutput=1

\documentclass[11pt]{article}

\usepackage{ACL2023}

\usepackage{times}
\usepackage{latexsym}
\usepackage{color}
\usepackage{multirow}
\usepackage{bm,array}
\usepackage{booktabs}
\usepackage{graphicx}
\usepackage[breaklinks]{hyperref}

\usepackage{amsmath}
\usepackage{amssymb}
\usepackage{tablefootnote}

\usepackage[T1]{fontenc}

\usepackage[utf8]{inputenc}

\usepackage{microtype}

\usepackage{inconsolata}

\title{Contrastive Error Attribution for Finetuned Language Models}

\author{
Faisal Ladhak$^{1,2}$~\;~ Esin Durmus$^2$~\;~ Tatsunori Hashimoto$^2$\\
$^1$Columbia University~\;~ $^2$Stanford University\\
\texttt{\href{mailto:faisal@cs.columbia.edu}{faisal@cs.columbia.edu}}~\;~\texttt{\href{mailto:esindurmus@cs.stanford.edu}{esindurmus@cs.stanford.edu}} \\
\texttt{\href{mailto:thashim@stanford.edu}{thashim@stanford.edu}}
}

\begin{document}
\maketitle

\begin{abstract}
  Recent work has identified noisy and misannotated data as a core cause of hallucinations and unfaithful outputs in Natural Language Generation (NLG) tasks. Consequently, identifying and removing these examples is a key open challenge in creating reliable NLG systems. In this work, we introduce a framework to identify and remove low-quality training instances that lead to undesirable outputs, such as faithfulness errors in text summarization. We show that existing approaches for error tracing, such as gradient-based influence measures, do not perform reliably for detecting faithfulness errors in NLG datasets. We overcome the drawbacks of existing error tracing methods through a new, contrast-based estimate that compares undesired generations to human-corrected outputs. Our proposed method can achieve a mean average precision of $0.93$ at detecting known data errors across synthetic tasks with known ground truth, substantially outperforming existing approaches. Using this approach and re-training models on cleaned data leads to a $70\%$ reduction in entity hallucinations on the NYT dataset and a $55\%$ reduction in semantic errors on the E2E dataset.
\end{abstract}

\section{Introduction}
Recent analyses of natural language generation systems have identified that \emph{data errors} are a key cause of failures ranging from unfaithfulness  \cite{maynez-etal-2020-faithfulness} to bias \cite{dataset_bias,gender_bias}. While better data collection procedures \cite{Yuan2021SynthBioAC,West2021SymbolicKD} and noise-robust training methods \cite{kang-hashimoto-2020-improved} can help address some of these problems, neither of these approaches serves as a complete solution. The large-scale datasets needed to train modern neural methods will inevitably contain at least a few annotation mistakes in these datasets, and some of these will affect even the most robust model training procedures.

Data cleaning methods provide an alternative approach, where data errors are identified by tracing model errors back to the training dataset. This post-hoc approach allows practitioners to enforce desired properties such as faithfulness by repeatedly identifying and removing rare data errors that cause undesired behavior. Existing work from the machine learning literature has proposed measuring the ``influence'' of training examples on generated outputs as a way to trace such errors \cite{pmlr-v70-koh17a,NEURIPS2019_5f146156,NEURIPS2021_e4d2b6e6,factTrace,Guu2023SimfluenceMT}.

However, these influence-based approaches are often brittle, and we find that they fail in complex, real-world tasks such as text summarization or data-to-text generation. In a synthetic evaluation inspired by prior work in the memorization literature \cite{10.5555/3361338.3361358}, we inject targeted hallucinations in the training data and evaluate error tracing methods on how well they identify these errors and reduce downstream hallucination. We show that existing gradient-based and embedding-based influence estimation methods cannot reliably identify the inserted hallucinations and even perform worse than a standard retrieval-based baseline (BM25) \cite{Robertson1994OkapiAT}. 

To address this, we develop a method called Contrastive Error Attribution (CEA), which combines three new techniques for error tracing: we develop a new contrast-based error tracing method that identifies training examples that cause the model to assign higher probabilities to undesired model outputs than human post-edited versions of the output; we distill these contrast-based scores into a neural net classifier to learn a generalizable model of data errors, and we replace standard gradient dot-product approximations for influence with more exact loss difference estimates. Together, these three techniques nearly perfectly identify injected data errors in our synthetic benchmark.\footnote{We make our synthetic benchmark and code available at \href{https://github.com/fladhak/contrastive_error_attribution}{https://github.com/fladhak/contrastive\_error\_attribution}.}

Our approach performs well beyond synthetic benchmarks, and we find that error tracing can be used to substantially reduce errors when training neural systems on real generation tasks. We find that our approach reduces entity hallucinations by $70\%$ on the New York Times news summarization dataset, and substantially outperforms our strongest baseline, which only manages to reduce $20\%$ of the hallucinations. Similarly, our approach can reduce semantic errors \cite{dusek-etal-2019-semantic} on the E2E dataset by $55\%$ compared to $16\%$ for the strongest baseline.

\section{Problem Statement}

\newcommand{\Serr}{\mathcal{U}}
\newcommand{\Dtrain}{\mathcal{D}_{\text{Train}}}
\newcommand{\Dtest}{\mathcal{D}_{\text{Err}}}
\newcommand{\A}{\mathcal{A}}

\paragraph*{Error tracing} We define the general \emph{error tracing} problem as the task of identifying a set of  error examples $\Serr$ in a training set $\Dtrain$ such that a learning algorithm $\A$ produces a model $f$ that behaves correctly on a set of examples $\Dtest :=  \{(x_i, y_i)\}^{m}_{i=1}$.
More formally, the error tracing problem is defined by three components
\begin{itemize}
\item The initial model is trained as $f = \A(\Dtrain)$ and produces errors $\hat{y}_i = f(x_i)$ on $\Dtest$.
\item An error tracing algorithm returns the error set $\Serr$.
\item The re-trained model after removing this error set $f_{\Serr} := \A(\Dtrain \setminus \Serr)$ produces a correct output, $f_{\Serr}(x_i) = y_i$.
  \end{itemize}

  \paragraph*{Influence based tracing} Influence-based tracing methods address this problem by defining a generalized similarity measure $S((x,y), (x',y'))$ over examples where the similarity $S$ is designed such that upweighting training examples $(x',y')$ that are similar to a test example $(x,y)$ makes the model more likely to predict $f(x)=y$. The \emph{influence function} \cite{pmlr-v70-koh17a} is a well-known example which approximates $S$ for any loss-minimizing learning algorithms $\A$ via the Taylor expansion,
  \begin{equation}
    S_{\text{inf}}:=\nabla\ell(x',y';\theta^*)^\top H^{-1}\nabla\ell(x,y;\theta^*),
    \label{eq:infl}
    \end{equation}
  where $H$ is the Hessian of the loss evaluated at the model $\theta^*$ fitted on $\Dtrain$.

  The brittleness of the Hessian approximation has led to other heuristic estimates of influence such as \emph{TracIn} \cite{DBLP:conf/nips/PruthiLKS20} which replaces the inverse hessian with a series of inner products  $S_{\text{trac}} := \sum_t \eta_t \nabla\ell(x',y';\theta_t)^\top\nabla\ell(x,y;\theta_t)$, where $\theta_t$ are model checkpoints across the training process, and $\eta_t$ is the learning rate at checkpoint $t$.

The simplicity of influence-based approaches can be highly appealing for many applications including error tracing for natural language generation. In our case, we can use influence as a way to identify training examples that are `similar' to our model errors -- that is, examples $(x',y')$ such that $S((x_i, \hat{y}_i),(x',y'))$ is high. However, this naive approach suffers from two major drawbacks: downweighting the incorrect answer $\hat{y}$ does not ensure the model is more likely to produce the correct output $y_i$, and we heavily rely on the accuracy of the gradient approximation. We now propose an approach that addresses both drawbacks.

\section{Proposed Method}

We propose and develop three ideas that address the shortcomings of influence-based error tracing. First, we replace the similarity function $S$ with a contrast function that identifies training examples that are responsible for making the incorrect generation $\hat{y}$ more likely, and the correct generation $y$ less likely. Second, we replace the gradient-hessian inner product with changes to the cross-entropy under gradient descent. Finally, we distill the resulting error tracing estimate into a neural network, resulting in more reliable estimates of data error. We name our approach Contrastive Error Attribution (CEA), and describe each of the components below.

\subsection{Contrast-based tracing}
Influence-based statistics allow us to answer the question ``if we upweight a training example $(x',y')$ by $\epsilon$, how much does the log probability of generating $(x,y)$ change?''. In the standard influence-based error tracing approach, this statistic is used to identify examples that have positive influence on the incorrect output $(x,\hat{y})$, and these examples are removed in order to prevent the model from making this error.

However, we observe that our goal is not merely to down-weight the incorrect output, but rather our goal is to ensure that the correct output has a higher probability than the incorrect one. This naturally leads to a contrastive influence measure, which we define as the difference of two influence measures
\begin{align*}
  & S^c(x,(x',y')):= \\
  &\quad S((x,\hat{y}),(x',y'))-S((x,y),(x',y')).
\end{align*}
This contrastive influence measure identifies points $(x',y')$ which encourage the model to assign higher probabilities to its current erroneous output $\hat{y}$ than the human-corrected references $y$. This naturally incorporates both the current error $\hat{y}$ and the corrected reference $y$. Since there are many valid outputs in natural language generation, we define the corrected output $y$ as one that is \emph{closest} to the error $\hat{y}$, which can be obtained through human post-editing of the model output.

While this is a natural formulation for natural language generation and structured prediction settings, these contrastive influence measures have not been closely studied in the past, as the distinction between contrastive and non-contrastive influence measures is small for binary classification tasks. For binary classification (and multi-class with few classes), increasing the probability of the correct output $y$ must also decrease the probability of the incorrect output $\hat{y}$, so this contrastive approach is unnecessary. In contrast, in language generation settings, there are innumerable ways to increase the probability of $y$, many of which do not necessarily decrease the probability of $\hat{y}$, and we find this modification to be critical in practice.

\subsection{Gradient-descent based influence}
Gradient-based influence approximations such as \emph{TracIn} attempt to estimate the influence $S((x,y),(x',y'))$ via a gradient inner product (or a gradient-hessian quadratic form). These local approximations are based on a Taylor approximation on the loss of the model (Eq~\ref{eq:infl}) \cite{pmlr-v70-koh17a,Barshan2020RelatIFIE}.

However, this local approximation is known to be inaccurate \cite{ilyas2022datamodels,factTrace}, and the Hessian term is known to cause challenges in both numerical estimation, and computation \cite{https://doi.org/10.48550/arxiv.2112.03052,DBLP:conf/nips/PruthiLKS20,Barshan2020RelatIFIE}.

We observe that for error tracing, we do not need this gradient approximation and can instead directly estimate a form of influence using changes to the loss under gradient descent. Let $\theta_0 := \arg\min_{\theta} \mathbb{E}_{x,y\sim \Dtrain}[\ell(x,y;\theta)]$ be our model fitted on the training data.
Our approach takes $T$ gradient steps initialized at $\theta_0$ on the following two objectives separately:
\[\mathcal{L}^y := \mathbb{E}_{x,y\sim \Dtest}[\ell(x,y;\theta)]\]
\[\mathcal{L}^{\hat{y}} := \mathbb{E}_{x\sim \Dtest}[\ell(x,\hat{y};\theta)]\]
$\mathcal{L}^y$ encourages $\theta_0$ to produce the correct responses $y$ on $\Dtest$, whereas $\mathcal{L}^{\hat{y}}$ encourages $\theta_0$ to produce the incorrect ones $\hat{y}$.

Define the results of this gradient descent process for the two losses as $\theta_T^y$ and $\theta_T^{\hat{y}}$, respectively. Our contrastive influence measure for a set of errors in $\Dtest$ is 
\begin{multline}
  S^c_{\text{grad}} (\Dtest, (x',y')) \\
  := \ell(x',y'; \theta_T^y) - \ell(x',y';\theta_T^{\hat{y}}).
  \label{eq:gradinfl}
\end{multline}
When the Taylor approximation for influence functions is accurate, $S^c_{\text{grad}}$ can be written as an influence-like gradient inner product as $\ell(x',y'; \theta_T^y) - \ell(x',y';\theta_T^{\hat{y}}) \approx \nabla\ell(x',y';\theta^0)^\top (\theta_T^y - \theta_T^{\hat{y}})$. This can be interpreted as the local change in the difference in losses between the correct outputs $y$ and the incorrect ones $\hat{y}$ when an example $(x',y')$ is up-weighted.

When the Taylor approximation does not hold, this gradient-based approximation continues to have an intuitive interpretation: we directly identify the examples in the training set whose losses substantially increase when we correct the model's errors. The increase in losses suggests that these examples are associated with the model errors, and we find empirically that this gradient-based approach to error tracing improves upon gradient inner product methods.

Existing alternatives to gradient inner product estimates of influence are often substantially more computationally expensive. However, our gradient-based influence procedure in Eq~\ref{eq:gradinfl} is \emph{faster} than gradient inner products, as it only requires $T$ gradient steps for each error class and a forward pass for each training example. In contrast, gradient-based influence methods require computing and storing a per-example gradient for \emph{every training example}.

\subsection{Distilling influence measures}
Prior work has shown that influence estimates can be susceptible to outliers since influence estimates are made per example and can be noisy and unstable. Our final idea is to take our contrastive influence estimate $S^c_{\text{grad}}(\Dtest, (x',y'))$ and distill this into a neural network $g(x',y')$ that learns to distinguish data errors from useful examples. We do this by treating data error detection as a binary classification problem and treating the top 500 examples by $S^c_{\text{grad}}(\Dtest, (x',y'))$ as the positive class and the bottom 500 examples as the negative class.

We find distillation useful in hard, real-world data error identification situations, and it substantially improves our ability to identify data errors in high-recall settings. Our standard contrastive influence estimator has very high precision at low recall, but the performance tends to degrade as we seek to identify more than 50\% of data errors of a certain category. Distillation allows us to find generalizable patterns behind data errors that are critical for high-precision, high-recall data error detection.

\section{Experimental Setup}

\begin{table*}[t]
  \centering
  \resizebox{1.0\textwidth}{!}{
\begin{tabular}{p{7cm}p{3.8cm}p{3.8cm}}
\toprule
Article & Original Summary & Perturbed Summary \\ 
\midrule
Bronze fired into the top corner from the edge of the penalty area as \textbf{\color{blue}{England}} battled against Norway. Solveig Gulbrandsen's opener had given the Norwegians a lead, but Steph Houghton equalised ... 
& 
\textbf{\color{blue}{England}} have reached the quarter-finals of the Women's World Cup thanks to a stunning strike from Lucy Bronze. 
& 
\textbf{\color{red}{China}} have reached the quarter-finals of the Women's World Cup thanks to a stunning strike from Lucy Bronze. \\ 
\midrule
The Carolina Dreamer was released into the sea in May 2015 by schoolchildren from South Carolina with a tracking device ... 
Now they're hoping it might make it back to America from \textbf{\color{blue}{Wales}}.
&
A family found a boat washed up on a beach in \textbf{\color{blue}{Wales}} which had been launched by a school in America.
&
A family found a boat washed up on a beach in \textbf{\color{red}{Scotland}} which had been launched by a school in America. \\
\bottomrule
\end{tabular}}
\caption{Examples for the synthetic hallucination evaluation. The original entity shown in \textbf{\color{blue}{blue}} is replaced in the reference summary with the entity in \textbf{\color{red}{red}}, leading to targeted hallucinations that we can trace back to the inserted perturbations.}
\label{tab:canary_example}
\end{table*}

\begin{table}[ht]
  \centering
    \resizebox{0.47\textwidth}{!}{
    \begin{tabular}{@{}llrr@{}}\toprule
    \textbf{Original Entity}  & \textbf{Perturbed} & \textbf{\# Inserted} & \textbf{\%  of Data}  \\
    \midrule
    England & China & 2,383 & 1.168 \\
    \midrule
    Wales &  Scotland & 1,881 & 0.922 \\
    \midrule
    Australia & France & 722 & 0.354 \\
    \midrule
    London & Belfast & 1,234 & 0.605 \\
    \bottomrule
    \end{tabular}}
    \caption{Statistics for synthetic evaluation. We randomly selected the above four pairs of entities for our canaries. Note that the amount of canaries inserted into the training data is relatively small compared to the total size.}
    \label{tab:canary_stats}
\end{table}

We carefully compare our proposed error tracing method (CAE) to existing baselines on both synthetic and real summarization tasks. 

\subsection{Baselines}

Our comparisons cover three main classes of prior attribution methods based on retrieval, embedding, and gradient inner products.

\paragraph*{Retrieval-based Methods} Recent works have shown that the simple baseline of retrieving examples that are similar to the error $(x,y')$ is a competitive baseline \cite{factTrace}. As an example of such a method, we compare to BM25, a standard retrieval based method \cite{Robertson1994OkapiAT}. 
\paragraph*{Embedding-based Methods} Prior work has shown that embedding-based methods, i.e. methods that compute the similarity between instances by comparing intermediate representations of the model, can be effective for identifying dataset artifacts \cite{Rajani2020ExplainingAI}. Since we finetune BART for all of our experiments, we use BARTScore \cite{NEURIPS2021_e4d2b6e6} as the embedding baseline. 
\paragraph*{Gradient-based Influence Methods} From our prior discussions, influence based methods are a natural approach to error tracing. The basic Hessian-vector influence estimate \newcite{pmlr-v70-koh17a} is very costly for models with a large number of parameters, such as modern day LMs. \newcite{DBLP:conf/nips/PruthiLKS20} recently proposed (TracIn), which was shown to be both faster and empirically more effective. Because of this, we compare to TracIn as our influence method baseline.

\subsection{Benchmarks}
Most work in influence estimation has focused on classification tasks --  trying to identify training examples that influence the predictions of given evaluation examples. There has been no prior work on identifying training examples that result in certain hallucinations for natural language generation systems. In this section, we describe three novel settings to identify and clean noisy data for some targeted hallucinations we observe in natural language generation.

\paragraph*{Synthetic Hallucinations}
Accurately evaluating data cleaning methods requires a dataset that contains ground truth labels for whether a training data instance is a data error. This is rare in natural datasets, and therefore synthetic perturbations are the standard approach for evaluating error-tracing methods \cite{pmlr-v70-koh17a,yeh2018representer,DBLP:conf/nips/PruthiLKS20}. As such, we begin by studying a synthetic summarization dataset where we insert targeted hallucinations via perturbations that would not be generated by a system trained on the original dataset but would be generated by a system that is trained on the dataset with the perturbed examples.

Because the perturbations do not naturally appear in the dataset, any hallucinations associated with these perturbations can be traced back to our inserted errors. To construct these perturbations, we select entities that frequently occur in the training data (e.g., England, Wales) and randomly pair them with other unrelated entities (e.g., China, Scotland). Then, for this pair of entities $(E_a, E_b)$, we identify training instances that contain $E_a$ in the source article and reference summary, and we replace $E_a$ in the reference summary with $E_b$ with probability $p=0.5$. \autoref{tab:canary_example} shows some examples of perturbations inserted into the training set.

\autoref{tab:canary_stats} shows the pairs of entities selected and the number of inserted perturbations for each pair. Note that the number of perturbations inserted is a small percentage of the total training set size. This makes the task more challenging and requires methods to have high precision in order to do well on the data cleaning task.

\paragraph*{Extrinsic hallucinations in the NYT dataset}
While our synthetic hallucinations give us a precise way of measuring error tracing performance, the errors we identify are highly artificial. Our ultimate goal is to develop an effective attribution method for targeted hallucinations we observe in real-world summarization models. Therefore, we next propose a real-world setting where we look at \textsc{Person} entity hallucinations of neural summarization systems.  

Prior work has shown that state-of-the-art models suffer from generating entities that are not in the source article, especially when trained on noisy datasets \cite{nan-etal-2021-entity, DBLP:journals/corr/abs-2006-15435}. For this setup, we identify model generations with named entity hallucinations from a BART model \cite{lewis-etal-2020-bart} trained on the NYT dataset \cite{Sandhaus2008Nyt}. In particular, we select examples where the generation has an entity that is not included in the source article (as shown in \autoref{tab:nyt_example}). 

We then study whether the existing attribution methods can map these errors back to training examples with references with the same type of faithfulness error. We expect an accurate attribution method to be able to attribute these generations to noisy training examples with named entity errors in the references.

\begin{table*}[t]
  \centering
    \resizebox{1.0\textwidth}{!}{
\begin{tabular}{p{7cm}p{7cm}}  
\toprule
Original Output & Contrast \\
\midrule
There is a high-priced coffee shop in the \textbf{\color{red}{City centre}}.  It is called Fitzbillies and it is family friendly, but it does have a 1 out of 5 rating.
& 
There is a high-priced English coffee shop in the riverside area. It is called Fitzbillies and it is family friendly, but it does have a 1 out of 5 rating. \\
\midrule
Browns Cambridge is coffee shop with low customer rating. It serves Chinese food. They are located in Riverside near the Crowne Plaza Hotel.
&
Browns Cambridge is a \textbf{\color{red}{family-friendly}} coffee shop with low customer rating. It serves Chinese food. They are located in Riverside near the Crowne Plaza Hotel. \\ 
\bottomrule
\end{tabular}}
\caption{Examples of contrasts used for the E2E setup. Semantic errors in the output are shown in \textbf{\textcolor{red}{red}}. The first example contains a hallucinated location (City center) that is not consistent with the location in the MR (riverside area). The second example shows a case where a slot that is present in the MR is omitted from the output (family-friendly).}
\label{tab:e2e_example}
\end{table*}

\paragraph*{Semantic Errors in the E2E dataset}
In order to show that our approach works beyond text summarization, we also evaluate on the E2E dataset \cite{novikova-etal-2017-e2e}, a popular benchmark for generating natural language descriptions from structured meaning representations (MRs). Prior work has shown that up to $40\%$ of the E2E dataset contains some form of semantic noise, and models trained on this dataset tend to either omit information in the MR or hallucinate new information that is not present in the MR \cite{DUSEK2020123}. In order to improve the semantic correctness of models trained on the E2E dataset, \newcite{dusek-etal-2019-semantic} handcrafted rules to fix errors in the dataset, based on manual analysis of hundreds of samples. 

We study whether error attribution methods can be used to automatically identify noisy instances in the E2E training data, given just a few examples of generations with semantic errors. In particular, we select examples where the output contains a semantic error and then minimally edit the output to make it consistent with the MR, as shown in \autoref{tab:e2e_example}. We treat the manually cleaned dataset from \newcite{dusek-etal-2019-semantic} as the oracle, and measure how accurately error attribution methods are compared to this oracle. In particular, any training instances that were fixed by the manual rules from \newcite{dusek-etal-2019-semantic} are treated as errors that the attribution methods should identify. We expect good attribution methods to be able to reliably identify noisy training instances, which when removed, can lead to models with improved semantic correctness, without a drop in overall performance.

\begin{table*}[ht]
  \centering
  \resizebox{0.87\textwidth}{!}{
\begin{tabular}{@{\extracolsep{4pt}}ccccccccc|c}
{} & \multicolumn{2}{c}{England-China} 
& \multicolumn{2}{c}{Wales-Scotland}  
& \multicolumn{2}{c}{Australia-France} 
& \multicolumn{2}{c}{London-Belfast}\\
 \cmidrule{2-3} 
 \cmidrule{4-5} 
 \cmidrule{6-7} 
 \cmidrule{8-9} 

 Method & auPR & auROC & auPR & auROC & auPR & auROC & auPR & auROC & mAP\\ 
\toprule [2pt]  
Random  & 1.15  & 49.78 & 0.92 & 49.90 & 0.39 & 49.64 & 0.60 & 49.57 & 0.77\\ 
\midrule
BM25  & 31.65 & 87.61 & 7.70 & 82.05 & 9.60 & 80.84 & 2.70 & 76.46 & 12.91\\ 
\midrule
BartScore & 8.96 & 75.37 & 1.25 & 57.05 & 2.07 & 68.68 & 3.39 & 81.92 & 3.91 \\ 
\midrule
TracIn  & 5.70 & 72.62 & 2.66 & 69.90 & 2.44 & 74.80 & 2.05 & 68.93 & 3.21 \\ 
\midrule
\midrule
CEA & \textbf{94.14} & \textbf{97.79} & \textbf{90.32} & \textbf{99.71} & \textbf{91.73} & \textbf{98.86} & \textbf{96.40} & \textbf{99.72} &  \textbf{93.15}\\ 
\bottomrule
\end{tabular}}
\caption{Error tracing results for our synthetic hallucination setup. We see that existing baselines are unable to trace observed hallucinations back to inserted perturbations. Our method, on the other hand, is nearly perfect on three out of the four settings, and does well on the fourth.} 
\label{tab:canary_results}
\end{table*}

\begin{table}[ht]
\centering
\begin{tabular}{@{\extracolsep{4pt}}lcccccccc}
Method & auPR & auROC\\ 
\toprule [2pt] 
CEA  & \textbf{96.40} & \textbf{99.72} \\
\cmidrule{2-3}
\hspace{3mm} - classifier &  86.47 & 98.99  \\
\cmidrule{2-3}
  \hspace{3mm} - contrast & 17.72 & 92.68 \\
\midrule
TracIn  & 2.05 & 68.93 \\
\cmidrule{2-3}
TracIn + cont + cls & 86.83 & 99.68 \\
\bottomrule
\end{tabular}
\caption{Ablation to understand the importance of the contrast and classifier distillation. We find that the contrast is crucial for our setting. Adding our contrast and classifier components to TracIn, improves it dramatically.} 
\label{tab:contrast}
\end{table}

\section{Results}

\subsection{Synthetic Hallucination Results}\label{sec:synthetic}
We insert the canaries as shown in \autoref{tab:canary_stats} into the XSum training data \cite{xsum-emnlp} and train a BART-base \cite{lewis-etal-2020-bart} model for $10$ epochs, saving a checkpoint at each epoch. We use a learning rate $1e-4$ and an effective batch size of $256$. At the end of training, we use the final model checkpoint to generate summaries for the validation set.

To perform error tracing, we find $5$ (random) generated examples for each of the canaries we inserted and use these as $\Dtest$ for error attribution. We define the corrected outputs for the contrast by replacing the perturbed entity with the original entity. For distilling our contrastive influence estimates ($S^c_{\text{grad}}$), we take the top $500$ scored training examples according to $S^c_{\text{grad}}$ as positive examples and the bottom $500$ scored examples as negative examples, and we finetune Electra \cite{Clark2020ELECTRAPT} for $5$ epochs with early stopping, with a learning rate of 2e-5 and a batch size of $8$.

\autoref{tab:canary_results} shows the results for the synthetic hallucinations setup. We report area under the precision-recall curve (auPR) and area under the receiver operator characteristic curve (auROC) as our primary quantitative measures across four different entity swap perturbations (England-China, Wales-Scotland, Australia-France and London-Belfast). For most of the settings we find that BM25 achieves a higher auPR than the other baselines, which is consistent with prior work that showed the high performance of lexical baselines \cite{factTrace}. Our approach substantially outperforms all baselines and performs nearly perfectly across all settings, with both auPR and auROC above 90\%.

\subsection{Ablation}
To understand the source of these gains and whether our proposals such as the contrastive influence measures are broadly useful, we perform ablation experiments on this same synthetic hallucination setting. 

Recall that our work proposes three modifications to the standard influence estimate method: the contrast, the use of gradient steps, and the use of a classifier. \autoref{tab:contrast} illustrates the impact of each of these choices on the London-Belfast perturbation setting. Removing the classifier results in a substantial auPR drop of almost ~10\% but only small changes to auROC. Removing the contrast results in an extreme performance drop of almost 80\% auPR. Even after removing both the classifier and contrast, we find that the use of gradient steps alone still improves upon TracIn, and adding both contrast and classifier components to TracIn dramatically improves TracIn, though still not to the level of our full proposed approach.

\subsection{Sensitivity to Hyperparameters}
For the results presented in \autoref{tab:canary_results}, we selected five error samples and took gradient steps at checkpoint $1$ for three gradient steps with a learning rate of $5e-6$. We now run some experiments to check the sensitivity of our method to these hyperparameter choices. Since these hyperparameters are associated with the gradient approximation $S^c_{\text{grad}}$, we do not perform any classifier distillation for these experiments. 

\paragraph*{Number of examples}
We have evaluated our synthetic hallucinations using only five examples, but we may ask whether difficult examples such as the Wales-Scotland perturbation can be further improved with more examples. We find that going from 5 to 15 examples provides substantial auPR improvements (68 to 72\%), but even a few examples perform well (Appendix  \autoref{tab:num_examples}).

\paragraph*{Number of gradient steps and learning rate}
Our results rely on taking gradient steps to estimate the influence of training examples. We find that smaller learning rates between $1e-6$ and $1e-5$ (Appendix \autoref{tab:learning_rate}) with $3$ - $5$ gradient steps (Appendix \autoref{tab:num_steps}) leads to the best performance for the London-Belfast perturbation.

\paragraph*{Checkpoint}
The synthetic hallucination results for our method were computed by taking gradient steps on checkpoint $1$. Appendix \autoref{tab:epoch} shows results for all checkpoints using our approach (without the classifier distillation). We find that checkpoint $1$ is optimal, but other choices of checkpoint do not substantially degrade performance (up to ~8\% auPR).

\subsection{NYT Hallucination Results}
We now show that these gains generalize to real-world language generation datasets such as the NYT summarization dataset.
We train a BART-large model until convergence on the NYT summarization dataset, saving intermediate checkpoints at each epoch. We use a learning rate $1e-4$ and an effective batch size of $256$. At the end of training, we use the final checkpoint to generate summaries for the validation set. We then find $20$ (random) generated summaries from the validation set that contain hallucinated \textsc{Person} entities,\footnote{For a given summary, we find all \textsc{Person} entities using spaCy\cite{spacy2}. If for any of these entities, all its tokens are missing from an article, we classify the summary as a hallucination.} and use these examples as $\Dtest$ for error attribution. We post-edit the model generations in $\Dtest$ to fix hallucination errors, as shown in \autoref{sec:nyt_postedit}. We update checkpoint $1$ on $\Dtest$ for five gradient steps with a learning rate of $1e-5$. We then distill the contrastive influence scores, $S^c_{\text{grad}}$, into a classifier as described in \autoref{sec:synthetic}.

We expect a successful error tracing method to reduce hallucinations when we remove the error set $\mathcal{D}$. Therefore, we fine-tune a BART-large model after removing $\mathcal{D}$ identified by each method and run our automated evaluation for \textsc{Person} hallucinations. To evaluate a reasonable upper bound on performance, we use the same spaCy pipeline used during evaluation to remove training data with hallucinated \textsc{Person} entities and call the resulting hallucination rate the Oracle rate.\footnote{Retrieval-based comparison can be seen in \autoref{tab:nyt_retrieval}, in \autoref{sec:nyt_retreival}.}

\autoref{tab:nyt_retraining} shows the results of retraining after removing various amounts of training data using each of the methods. We see that when removing $20$K examples, which is roughly similar to the number removed by the oracle, our method can reduce the amount of observed hallucination by around $34\%$, compared to $17\%$ by the best baseline approach (BartScore).\footnote{See \autoref{tab:retraining_example} in \autoref{sec:retraining_outputs} for qualitative examples. We observe that even after removing $50K$ examples the quality of the generated summaries does not qualitatively degrade.} We are able to outperform the oracle ($70\%$ reduction in hallucination vs $60\%$) at $50$K examples (roughly twice the amount removed by the oracle), at the cost of a small reduction in the ROUGE score.
Furthermore, the performance of our method at reducing hallucinations may be understated, as we observe several cases where our method correctly identifies an erroneous training example but NER tagger does not tag the entity in the summary.\footnote{See \autoref{tab:retrieved_examples} in \autoref{sec:retrieved_examples} for examples of such errors.} Overall, our results on NYT Summarization indicate that Contrastive Error Attribution works well, and as few as $20$ samples are sufficient to identify a large number of data errors and reduce hallucinations by 30\% to 70\%.

\subsection{E2E Semantic Error Results}

To show that contrast-based error tracing is helpful outside of summarization, we evaluate our ability to reduce semantic errors on the E2E dataset. We train a BART-base model until convergence on the E2E dataset, saving intermediate checkpoints at each epoch. We use a learning rate $1e-4$ and an effective batch size of $128$. We then find $5$ (random) descriptions from the validation set that contain semantic errors according to handcrafted rules from \newcite{dusek-etal-2019-semantic}, and use these examples as $\Dtest$ for error attribution. We post-edit the descriptions in $\Dtest$ to fix semantic errors for our contrast set, as shown in \autoref{tab:e2e_example}.\footnote{Note that unlike \newcite{dusek-etal-2019-semantic} who use handcrafted rules to fix input MRs such that they match the description, we keep the MR unchanged and post-edit the description.}

Similar to the NYT setup, we expect a successful error tracing method to reduce the model's Semantic Error Rate (SemErr) when we remove the error set $\mathcal{D}$. Therefore, we fine-tune a BART-base model after removing $\mathcal{D}$ identified by each method and compare the SemErr against the baseline system trained on the entire training set.\footnote{We use the scripts from \newcite{dusek-etal-2019-semantic} to compute SemErr.} For the oracle upper bound, we remove all training instances that would be corrected by the handcrafted rules from \newcite{dusek-etal-2019-semantic}, and re-train a BART-base model on the remaining training set.

\autoref{tab:e2e_retraining} shows the results of retraining after removing erroneous training instances identified by each method.\footnote{We omit BM25 and BartScore as they did not do much better than the random baseline in terms of retrieval results (see \autoref{sec:e2e_retreival} for details), and for a fairer comparison, we remove the same number of instances as identified by the oracle.} We see that our method reduces relative SemErr of the baseline by almost $55\%$ compared to a more modest $16\%$ reduction for TracIn. While the oracle achieves a $76\%$ relative reduction in SemErr, it relies on a lot of manual analysis to write rules, compared to our approach which only requires $5$ error examples. Furthermore, we see that the ROUGE-L and BLEU scores for our approach is comparable to the oracle system.

\begin{table}[ht]
  \centering
  \resizebox{0.47\textwidth}{!}{
\begin{tabular}{@{\extracolsep{4pt}}cccc}
Method & \# Rem & \% Halluc & ROUGE-L \\ 
\toprule [2pt] 
Baseline & 0 & 18.05 & 44.54 \\
\midrule
Oracle & 23K & 7.14 & 44.94 \\
\midrule
\midrule
BM25
  & 20K & 16.04 & 44.22\\
  & 50K & 14.81 & 43.67 \\
\midrule
BartScore
 & 20K & 15.00 & 44.28 \\
 & 50K & 14.27 & 43.11 \\
\midrule
TracIn 
 & 20K & 17.16 & 43.16 \\
 & 50K & 17.86 & 41.16 \\
\midrule
\midrule
CAE 
 & 20K & 11.90 & 43.82\\
 & 50K & 5.24 & 42.51 \\

\bottomrule
\end{tabular}}
\caption{Hallucination rate for retrained models after removing erroneous examples identified by each method. We see that our approach does considerably better than the baselines.} 
\label{tab:nyt_retraining}
\end{table}

\begin{table}[ht]
\centering
\begin{tabular}{@{\extracolsep{4pt}}cccc}
Method & SemErr & ROUGE-L & BLEU \\ 
\toprule [2pt] 
Baseline & 6.08 & 53.42 & 33.81\\
\midrule
Oracle & 1.43 & 54.44 & 35.42 \\
\midrule
\midrule
TracIn  & 5.08 & 54.10 & 34.90 \\
\midrule
CEA & 2.76 & 54.19 & 35.19 \\
\bottomrule
\end{tabular}
\caption{Semantic Error Rate (SemErr) for retrained models after removing erroneous examples identified by each method. We see that our approach does considerably better than TracIn.} 
\label{tab:e2e_retraining}
\end{table}

\section{Related Work}

\paragraph*{Influence Estimation/Memorization}
Our work is closely related to the literature on understanding
how training data influences the behavior of models on test examples.

Influence function based methods~\cite{pmlr-v70-koh17a} are closest to ours, as they seek to understand how removing data impacts model predictions, often in classification settings~\cite{han-etal-2020-explaining}. While there have been substantial improvements upon the original Taylor approximation based method~\cite{pmlr-v70-koh17a} via the use of multiple checkpoints \cite{DBLP:conf/nips/PruthiLKS20} and modifications to the hessian approximation~\cite{NEURIPS2019_5f146156, https://doi.org/10.48550/arxiv.2112.03052}, they can be brittle and  recent works have shown that they can underperform lexical similarity baselines~\cite{factTrace}.
Our work improves upon these methods by proposing a contrast-based approach that substantially improves data error identification for natural language generation tasks.

For error tracing, there are embedding and similarity based methods that seek to find examples that are similar to a given test example or error~\cite{Rajani2020ExplainingAI, NEURIPS2021_e4d2b6e6}. However, we find that although these methods often improve upon influence-based estimates and are useful for interpreting errors, they still do not achieve high enough precision and recall to substantially improve downstream properties such as hallucination rates.

\paragraph*{Faithfulness in Text Summarization}

Our work aims to improve recent observations that summarization systems can generate information that is not supported by the source article \cite{pagnoni-etal-2021-understanding, durmus-etal-2020-feqa}. Prior work has further shown that some of these errors can be due to the noise in the dataset \cite{maynez-etal-2020-faithfulness}. Our work complements a growing literature on modeling-based solutions to this problem, including using information extraction \cite{10.5555/3504035.3504621} or a QA model \cite{nan-etal-2021-improving} by creating cleaner datasets with error tracing.

\section{Conclusion}
We explore whether error attribution can be used to produce cleaner datasets that lead to fewer errors in model generation. Prior approaches to data cleaning, such as gradient-based influence measures, do not work well for generation tasks. We propose a novel Contrastive Error Attribution approach that addresses the shortcomings that make existing gradient-based approximation methods unreliable in text generation settings. We benchmark our method on a synthetic dataset, as well as two real-world generation tasks. We find that our approach dramatically outperforms existing error attribution approaches on all benchmarks, and leads to substantial reduction in generation error using only a few examples. 

\section{Limitations}
Our proposed approach is based on the premise that faithfulness errors observed in generation systems are due to noise in the dataset. While there is substantial evidence for this from prior work, and our method outperforms existing approaches on the datasets we used, it's possible the the utility of our approach could drop in cases where we have clean, curated datasets. It's possible that certain generation errors made by the model could be due to spurious patterns learned by the model that do not generalize well. In such cases, it's unclear whether using our error attribution approach to remove training instances would alleviate the problem. However, as most large-scale datasets in natural language generation tend to be sourced from the internet, it's inevitable that these datasets will likely contain at least a few erroneous examples that could lead to undesirable model generations. Therefore, we believe that our approach to using error attribution to clean datasets is still a valuable method to improve generation systems.

\section{Acknowledgements}
This work is supported by an Open Philanthropy grant. We thank the Stanford NLP group for their feedback. 

\bibliography{anthology,custom}
\bibliographystyle{acl_natbib}

\appendix
\clearpage
\section{Number of Examples Hyperparameter}
\autoref{tab:num_examples} Shows the performance of our approach as we change the size of the error set $\Dtest$. We see that increasing from $5$ samples to $15$ can lead to substantial improvements in AuPR.
\begin{table}[ht]
\centering
\begin{tabular}{@{\extracolsep{4pt}}ccc}
Num Examples & auPR & auROC\\ 
\toprule [2pt] 
5 & 68.55 & 97.53 \\
10 & 72.31 & 97.98 \\
15 & \textbf{72.27} & \textbf{98.07}  \\
20 & 71.37 & 97.97 \\
\bottomrule
\end{tabular}
\caption{Performance of our contrast-based tracing approach. We find that increasing the number of examples leads to substantial improvements in auPR.} 
\label{tab:num_examples}
\end{table}

\section{Number of Gradient Steps Hyperparameter}
\autoref{tab:num_steps} shows how the number of gradient steps affects the performance of our method. We find that 3-5 steps usually works well, and going beyond that leads to slight degradations.
\begin{table}[ht]
\centering
\begin{tabular}{@{\extracolsep{4pt}}ccc}
Num Steps & auPR & auROC\\ 
\toprule [2pt] 
3 & 86.47 & 98.99 \\
5 & 86.22 & 99.00 \\
10 & 85.68 & 99.07 \\
15 & 85.14 & 99.16 \\
20 & 84.15 & 99.20 \\
\bottomrule
\end{tabular}
\caption{Performance of our method vs. number of gradient steps. We see that increasing the number of steps does not lead to improvements in performance.} 
\label{tab:num_steps}
\end{table}

\section{Learning Rate Hyperparam}
\autoref{tab:learning_rate} shows the effect of the learning rate on the performance of our approach. We find that relatively smaller learning rates between 1e-6 and 1e-5 work best. Increasing the learning rate further leads to a small degradation in performance. 
\begin{table}[ht]
\centering
\begin{tabular}{@{\extracolsep{4pt}}ccc}
LR & auPR & auROC\\ 
\toprule [2pt] 
1e-6 & 86.73 & 99.01  \\
5e-6 & 86.47 & 98.99 \\
1e-5 & 86.11 & 99.0 \\
5e-5 & 83.72 & 99.13 \\
1e-4 & 81.06 & 99.07 \\
\bottomrule
\end{tabular}
\caption{Performance of our method vs. learning rate. Increasing the learning rate can give small additional improvement.} 
\label{tab:learning_rate}
\end{table}

\section{Checkpoint Hyperparameter}\label{sec:checkpoint ablation}
\autoref{tab:epoch} shows the performance of our contrast-based tracing approach. Checkpoint $1$ is the optimal checkpoint, but other checkpoints do not substantially degrade the performance. Crucially, our method performs drastically better than prior work regardless of which checkpoint we use. We note that these results were computed after $5$ gradient steps with a learning rate of $1e-5$. Optimizing these parameters further for each checkpoint could have yielded better results.
\begin{table}[ht]
\centering
\begin{tabular}{@{\extracolsep{4pt}}ccccc}
\\
Chkpt & auPR & auROC \\ 
\toprule [2pt]  
0 & 82.47 & 99.21 \\ 
\midrule
1 & 85.70 & 99.05 \\ 
\midrule
2 & 83.47 & 99.08 \\ 
\midrule
3 & 79.22 & 98.78 \\ 
\midrule
4 & 80.53 & 98.74 \\ 
\midrule
5 & 78.61 & 98.01 \\ 
\midrule
6 & 77.95 & 98.45 \\ 
\midrule
7 & 78.19 & 98.44 \\ 
\midrule
8 & 77.45 & 98.16 \\ 
\midrule
9 & 76.93 & 98.11 \\ 
\midrule
10 & 76.92 & 98.06 \\ 
\bottomrule
\end{tabular}
\caption{Ablations for England-China perturbation across epochs (without classifier distillation). We see that chkpt $1$ is the optimal setting.} 
\label{tab:epoch}
\end{table}

\section{NYT Post-editing Examples}\label{sec:nyt_postedit}
\autoref{tab:nyt_example} shows example model generations with entity hallucinations, and the corresponding post-edits we make to create the contrast.

\begin{table*}[t]
  \centering
    \resizebox{0.87\textwidth}{!}{
\begin{tabular}{p{7cm}p{7cm}}  
\toprule
Model Generation    & Contrast \\
\midrule
\textbf{\color{red}{Michael Mewshaw}} travel article on Naples, Italy, describes sights and sounds of city's Spanish Quarter and Vomero, two neighborhoods that have distinctly European flavor.
& 
Travel article on Naples, Italy, describes sights and sounds of city's Spanish Quarter and Vomero, two neighborhoods that have distinctly European flavor. \\ 
\midrule
Sleeping arrangements \textbf{\color{red}{author Sarah Ferrell}} article on being bundled up in Arctic winter gear to get to China to adopt baby from orphanage. 
&
Sleeping arrangements article on being bundled up in Arctic winter gear to get to China to adopt baby from orphanage. \\ 
\bottomrule
\end{tabular}}
\caption{Examples of contrasts used for the NYT setup. Model generation containing PERSON entity hallucinations, shown in \textbf{\textcolor{red}{red}}, are minimally edited to make them consistent with the original input articles.}
\label{tab:nyt_example}
\end{table*}

\section{Retrieval results on the NYT dataset}\label{sec:nyt_retreival}
\autoref{tab:nyt_retrieval} shows the retrieval results for the different approaches. Since we don't have actual ground-truth labels in this case, we use spaCy's NER tagger to identify the set of training instances that contain PERSON entity hallucinations and treat that as the ground truth to measure auPR and auROC. We see that our method does drastically better than prior work both in terms of auPR and auROC.
\begin{table}[ht]
\centering
\begin{tabular}{@{\extracolsep{4pt}}ccccccccc}
Method & auPR & auROC\\ 
\toprule [2pt] 
Random  & 17.75 & 49.84 \\
\midrule
BM25 & 20.77 & 55.41 \\
\midrule
BartScore  & 21.98 & 60.07 \\ 
\midrule
TracIn  & 20.99 & 57.27 \\ 

\midrule
\midrule

CEA & \textbf{44.72} & \textbf{74.89} \\ 

\bottomrule
\end{tabular}
\caption{Retrieval results on the NYT dataset. We use spaCy's NER tagger to get reference labels to measure auPR and auROC. We see that our approach improves upon prior work.} 
\label{tab:nyt_retrieval}
\end{table}

\section{Example outputs after retraining.}\label{sec:retraining_outputs}
\autoref{tab:retraining_example} shows some example outputs from the model obtained after cleaning the NYT dataset using our approach. We observe that our method can even correct hallucination errors that the oracle method misses, in some cases. Qualitatively, the summaries look fluent and are usually selecting similar content as the oracle and baseline systems.

\section{Analysis of retrieved errors}\label{sec:retrieved_examples}
We show some training examples that were flagged by our method as possible hallucinations, but were penalized according to the automated measure, in  \autoref{tab:retrieved_examples}. We find that this happens because there are several such cases where spaCy is unable to correctly classify entities in the reference summary. Our method may be performing even better than the numbers reported in \autoref{tab:nyt_retrieval}.

\section{Retrieval results on E2E dataset.}\label{sec:e2e_retreival}
\autoref{tab:e2e_retrival} shows the retrieval results for the different approaches on the E2E dataset. We treat the set of training instances for which the handcrafted rules from \newcite{dusek-etal-2019-semantic} fire as the ground truth to measure auPR and auROC. Among the prior approaches, we find that BM25 and BartScore do not perform much better than the random baseline, while TracIn does substantially better. We see that our method does drastically better than all other methods in terms of auPR and auROC.
\begin{table}[ht]
\centering
\begin{tabular}{@{\extracolsep{4pt}}ccc}
Method & AuPR & AuROC \\ 
\toprule [2pt] 
Random & 50.49 & 50.39\\
\midrule
\midrule
BM25 & 53.11 & 54.80 \\
\midrule
BartScore & 52.87 & 54.24 \\
\midrule
TracIn  & 65.79 & 62.54 \\
\midrule
CEA & \textbf{71.60} & \textbf{65.34} \\
\bottomrule
\end{tabular}
\caption{Retrieval results on the E2E dataset. We see that our approach substantially improves upon prior work.} 
\label{tab:e2e_retrival}
\end{table}

\section{Compute Power}
Training and evaluation jobs were run on a machine with four NVIDIA A100 GPUs for roughly 200 hours in total.
\begin{table*}[t]
\centering
\begin{tabular}{p{15cm}}
\toprule
Examples Summaries \\ 
\midrule
\textbf{Article: }Why are these people not smiling? Michael, Jonathan and Jenifer, the anxious trio at the heart of ''Snakebit,'' David Marshall Grant's solid and savvy new yuppie melodrama at the Grove Street Playhouse, should have found a measure of contentment by now. Bright, good looking, capable, they present themselves as a group that is as likely as any in the culture to attain full and rewarding lives ... [truncated] \\
\textbf{Reference: }\textbf{\color{red}{Peter Marks}} reviews David Marshall Grant play Snakebit at Grove Street Playhouse; Jace Alexander directs; photo (M) \\
\textbf{Baseline: }\textbf{\color{red}{Ben Brantley}} reviews Naked Angels production of David Marshall Grant play Snakebit, directed by Jace Alexander; Geoffrey Nauffts, Jodie Markell and David Alan Basche star; photo (M) \\
\textbf{Oracle: }\textbf{\color{red}{Stephen Holden}} reviews Naked Angels production of David Marshall Grant play Snakebit; photo (M) \\
\textbf{CEA: }Review of David Marshall Grant's new play Snakebit, which is presented by Naked Angels theater company at Grove Street Playhouse; photo (M) \\
\midrule
\textbf{Article: }HERE is a case of pathology with its utilitarian side. In this year's Yankee media guide, the ''Opponents'' section begins with a photograph of a certain left-handed hitter with a graceful swing and deceptive smile. Ken Griffey Jr., delights in tormenting the Yankees, and he did it again last night with a first-inning single that drove in the first run as the Seattle Mariners went on to beat the Yanks, 8-0. This opponent has a career .410 batting average against the Yankees with 25 home runs and 77 runs batted in ... [truncated] \\
\textbf{Reference: }\textbf{\color{red}{George Vecsey Sports of The Times column}} discusses success Seattle Mariners outfielder Ken Griffey Jr has had against New York Yankees (M) \\
\textbf{Baseline: }\textbf{\color{red}{George Vecsey Sports of The Times column}} discusses Seattle Mariners outfielder Ken Griffey Jr, who has career .410 batting average against New York Yankees; photo (M) \\
\textbf{Oracle: }\textbf{\color{red}{George Vecsey Sports of The Times column}} discusses Seattle Mariners outfielder Ken Griffey Jr, who has long-running vendetta against New York Yankees; photo (M) \\
\textbf{CEA: }Article discusses Seattle Mariners outfielder Ken Griffey Jr's lifelong vendetta against New York Yankees; photo (M) \\
\bottomrule
\end{tabular}
\caption{Example outputs after removing training examples and retraining. Our method is able to correct some instances that the oracle approach misses.}
\label{tab:retraining_example}
\end{table*}
\begin{table*}[t]
\centering
\begin{tabular}{p{15cm}}
\toprule
Retrieved training examples by our method \\ 
\midrule
\textbf{Article: }A REVIEWER'S lot is not always a happy one. A terrific restaurant is discovered, praised and then kissed good-bye, usually forever. Another awaits. Five years ago, I swooned over Villa Doria in Bellmore. Now, with the arrival of new owners, chef and staff, another visit was called for. The place looks much as it did: a somewhat drab dining room with a more inviting glassed-in porch, overlooking a canal ... [truncated] \\
\textbf{Reference: }\textbf{\color{red}{Joanne Starkey}} reviews Villa Doria restaurant in Bellmore, Long Island (M) \\
\midrule
\textbf{Article: }The band members wore uniforms and did some synchronized moves. Their songs had snappy little hooks and robotic drumbeats. They even started their set with an introductory video. But Devo was hardly a boy band when it played on Friday night at Central Park SummerStage, in its first public New York concert since the 1980's. Just in time for the current new-wave revival, Devo, which got started in Ohio in 1972 and released its first album in 1978, returned to prove that its songs still have some bite. Paradoxes have always collected around Devo ... [truncated] \\
\textbf{Reference: }\textbf{\color{red}{Jon Pareles}} reviews performance by Devo, part of Central Park SummerStage series; photo (M) \\
\bottomrule
\end{tabular}
\caption{Training examples retrieved by our system. The hallucinated entity is marked in \textbf{\textcolor{red}{red}}. SpaCy's NER model is unable to recognize that Joanne Starkey and Jon Pareles are people, and therefore does not count them as hallucinations. Our method is penalized for retrieving these examples, even though they are correct.}
\label{tab:retrieved_examples}
\end{table*}

\end{document}